\documentclass[final]{cvpr}

\usepackage{amsmath,amsfonts,bm}

\def\eqref#1{equation~\ref{#1}}

\def\1{\bm{1}}

\DeclareMathAlphabet{\mathsfit}{\encodingdefault}{\sfdefault}{m}{sl}
\SetMathAlphabet{\mathsfit}{bold}{\encodingdefault}{\sfdefault}{bx}{n}

\usepackage{times}
\usepackage{epsfig}
\usepackage{graphicx}
\usepackage{amsmath}
\usepackage{amssymb}
\usepackage{algorithm} 
\usepackage{algorithmic} 
\usepackage{multirow} 
\usepackage{amssymb} 
\usepackage{xcolor}
\usepackage{booktabs} 
\usepackage{subfigure}	
\usepackage{bbding}
\usepackage{pifont}
\usepackage{wasysym}
\usepackage{bm}

\usepackage[pagebackref=true,breaklinks=true,colorlinks,bookmarks=false]{hyperref}

\def\cvprPaperID{4002} 
\def\confYear{CVPR 2021}

\begin{document}

\title{Learning the Superpixel in a Non-iterative and Lifelong Manner}

\author{Lei Zhu$^{1,3,4}$ \\
\and Qi She$^{2}$ \\
\and Bin Zhang$^{4}$ \\
\and Yanye Lu$^{1,3,5}$ \\
\and Zhilin Lu$^{2}$ \\
\and Duo Li$^{2}$ \\
\and Jie Hu$^{2}$ \\
\and \\
\and $^{1}$Institute of Medical Technology, Peking University Health Science Center, Peking University \\
\and $^{2}$Bytedance AI Lab \\
\and $^{3}$Department of Biomedical Engineering, Peking University \\
\and $^{4}$Beijing University of Posts and Telecommunications \\
\and $^{5}$Institute of Biomedical Engineering, Peking University Shenzhen Graduate School \\
}

\maketitle

\pagestyle{empty}  
\thispagestyle{empty} 

\begin{abstract}
Superpixel is generated by automatically clustering pixels in an image into hundreds of compact partitions, which is widely used to perceive the object contours for its excellent contour adherence. Although some works use the Convolution Neural Network (CNN) to generate high-quality superpixel, we challenge the design principles of these networks, specifically for their dependence on manual labels and excess computation resources, which limits their flexibility compared with the traditional unsupervised segmentation methods. We target at redefining the CNN-based superpixel segmentation as a lifelong clustering task and propose an unsupervised CNN-based method called LNS-Net. The LNS-Net can learn superpixel in a non-iterative and lifelong manner without any manual labels. Specifically, a lightweight feature embedder is proposed for LNS-Net to efficiently generate the cluster-friendly features. With those features, seed nodes can be automatically assigned to cluster pixels in a non-iterative way. Additionally, our LNS-Net can adapt the sequentially lifelong learning by rescaling the gradient of weight based on both channel and spatial context to avoid overfitting. Experiments show that the proposed LNS-Net achieves significantly better performance on three benchmarks with nearly ten times lower complexity compared with other state-of-the-art methods. code is available at \url{https://github.com/zh460045050/LNSNet}. 
\end{abstract}
\section{Introduction}
Superpixel segmentation aims to over-segment an image into hundreds of compact partitions, \emph{i.e.} superpixel, by clustering the pixels based on both low-level color features and spatial features. Benefiting from concerning the spatial cues, the superpixel can be efficiently generated with high contour adherence. Therefore, it is widely used by both traditional machine learning (ML) and convolution neural network (CNN) to reduce computational complexity or perceive the contours of objects\cite{Seg-Track,DPO,DPO2,OCC}.

Many superpixel segmentation methods arise in the last decade including the gradient-based\cite{SLIC,SNIC,MSLIC,IMSLIC,SEEDS} and the graph-based methods\cite{ERS, DRW1,DRW2, LSC}. The gradient-based methods iteratively cluster the pixels in RGB or LAB space with limited spatial distance to refine the initialized cluster centers. This type of method has high efficiency, but suffers from low adherence due to their insufficient features. On the other hand, the graph-based algorithms usually have high adherence because they enrich the features by constructing an undirected graph. Afterwards, the subgraphs are generated as superpixel by cutting or adding edges to optimize a target energy function, which costs a lot of time.
\begin{figure}
\centering
\includegraphics[width=0.45\textwidth]{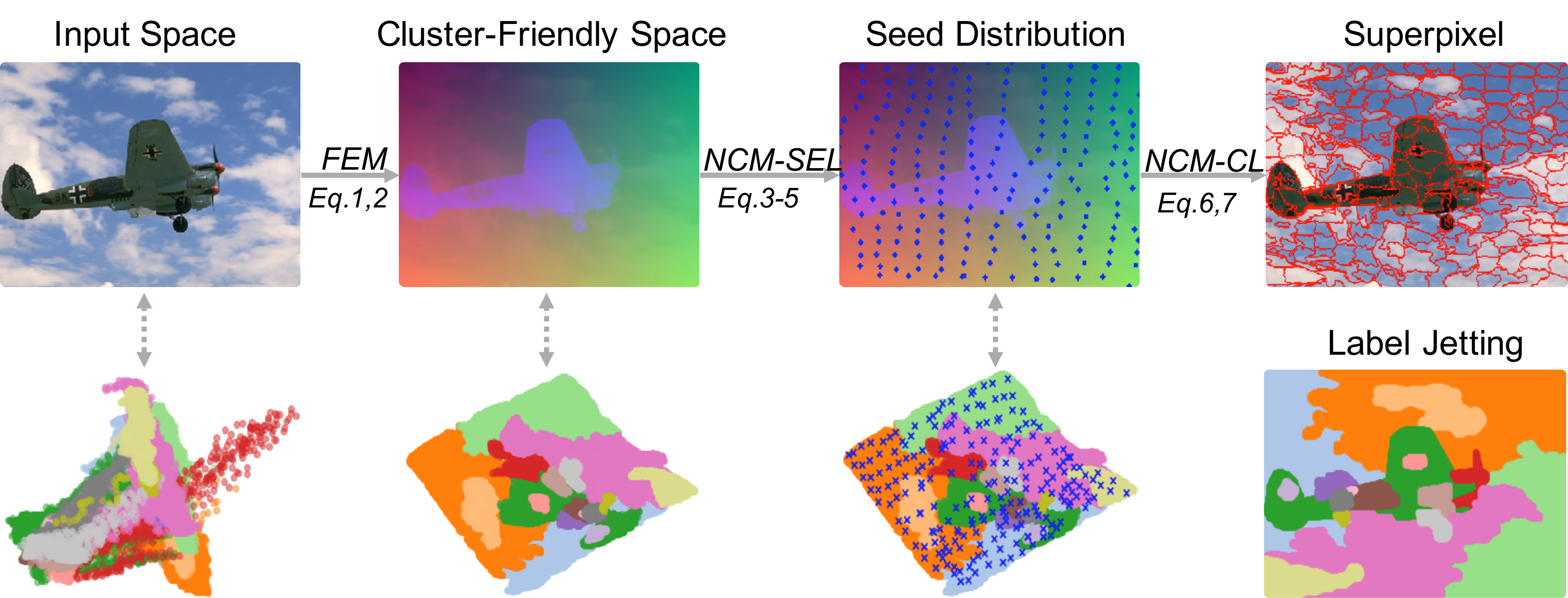}
\caption{The illustration of the workflow for the proposed LNS-Net, where the top row is the visualization of the features and the bottom is the distribution of the labels jetting in the feature space. The blue "x" is the generated seed node.}
\label{fig:one}
\end{figure}

Recently, benefiting from the prosperity of the CNN, some approaches employ the CNN to learn a suitable embedding space for superpixel segmentation and then cluster the pixels in this new feature space with clustering methods\cite{SSN, RIM, S-FCN, SEAL}. Even though they improve the performance by a large margin, some problems come into being simultaneously. Firstly, majority of the CNN-based methods\cite{S-FCN, SSN, SEAL} need human-labeled ground truth to supervise the network training which requires additional human resources to label all the pixels in images. Secondly, their offline training step needs to store all the training samples, which demands large amounts of memory and limits their flexibility to transfer the network into other domains. Finally, some CNN-based methods still need to iteratively update the coarse cluster centers (usually the center position of each grids), which is inconvenient and time-consuming.

To solve these problems, we redefine the CNN-based superpixel segmentation as a lifelong learning task\cite{CL1, CL2, CL3} which can sequentially learn a unified model online. In addition, a lightweight unsupervised CNN-based superpixel segmentation method called LNS-Net is proposed to learn superpixel in a non-iterative and lifelong manner. The LNS-Net is composed of three parts: \emph{feature embedder module} (FEM), \emph{non-iterative clustering module} (NCM) and \emph{gradient rescaling module} (GRM). Specifically, in the forward step shown in Fig.~\ref{fig:one}, FEM firstly embeds the original feature into a cluster-friendly space to protect detail cues with low complexity. Based on the cluster-friendly features, the proposed NCM uses a seed estimation layer (SEL) to learn the spatial shift of the central position, which directly estimates the optimal cluster centers, \emph{i.e.} the seed nodes. Then, the superpixel can be non-iteratively generated by the cluster layer (CL) of NCM that assigns the cluster for each pixel based on their similarity with the feature of seed nodes. Moreover, the GRM is proposed to solve the catastrophic forgetting caused by lifelong learning during backward step. It is consisted of gradient adaptive layer (GAL) and gradient bi-direction layer (GBL), which are used to avoid over-fitting by rescaling the gradient of each weight parameter based on channel importance and spatial context. A range-limited cluster loss is also proposed to effectively train our network without any manual labels.

In a nutshell, our main contributions are threefold: 1) To our knowledge, our work is the first to define the superpixel segmentation as lifelong learning task theoretically and give a corresponding solution. 2) A lightweight LNS-Net is proposed to non-iteratively generate the superpixel, which can be lifelong trained without any manual label. 3)  Experiments show that our LNS-Net has higher performance than other unsupervised methods and is also comparable with the supervised CNN-based methods.

\section{Related Works}

\noindent \textbf{Traditional Superpixel Methods:} The traditional superpixel segmentation methods include the gradient-based methods and the graph-based methods. The former iteratively cluster the pixels with limited spatial distance solely based on their color feature. 
Achanta \emph{et al.} proposed the simple linear iteratively clustering (SLIC)\cite{SLIC} to efficiently generate superpixel by limiting the search range of k-means. 
To further improve the efficiency, Achanta \emph{et al.} subsequently enabled the method to update cluster center and arrange the label of pixels simultaneously by proposing the simple non-linear iteratively clustering (SNIC)\cite{SNIC}.
Liu \emph{et al.} designed the manifold simple linear iteratively clustering (MSLIC)\cite{MSLIC,IMSLIC}, which adopts an adaptive search range for SLIC. 
Shen \emph{et al.} utilized another robuster cluster method called the density-based spatial clustering with noise (DBSCN)\cite{DBSCN} to generate superpixel with stronger spatial consideration. 
Different with the gradient-based methods, the graph-based methods firstly construct an undirected graph based on the feature of input image and then generate superpixel by creating sub-graphs. 
Shen \emph{et al.} proposed the lazy random walk (LRW)\cite{LRW}, which adds a self-loop into the random walk graph to make the walking process lazy and can be extended into the superpixel segmentation with the help of a shape-concerned energy term. 
Liu \emph{et al.} elaborated an entropy rate superpixel (ERS)\cite{ERS} that maximizes the random walk entropy by continually adding edges into the graph model. 
Li \emph{et al.} proposed the linear spectral superpixel clustering (LSC)\cite{LSC} to approximate the normalized cut (NCut)\cite{NCut} energy by weighted k-means cluster. 
Recently, Kang \emph{et al.} designed the dynamic random walk (DRW)\cite{DRW1, DRW2}, which efficiently improves the adherence of superpixel by proposing a weighted random walk entropy with limited walk range.

\noindent \textbf{CNN-based Superpixel Methods:} The CNN-based superpixel segmentation methods use the CNN to extract features and then cluster the pixels based on these features. Tu \emph{et al.} firstly adopted the CNN in superpixel segmentation by proposing a segmentation-aware loss (SEAL)\cite{SEAL}. It uses the ground truths of semantic segmentation (or boundary detection) to supervise the feature learning. However, SEAL cannot generate superpixel in an end-to-end mode because it adopts the time-consuming ERS\cite{ERS} as post-processing.
Jampani \emph{et al.} proposed an end-to-end superpixel segmentation network called superpixel sample network (SSN)\cite{SSN} by integrateing SLIC. SSN can be easily used to assist other vision tasks such as semantic segmentation with the task-specific loss. But, it still needs manual labels to supervise the network training and requires iteratively updating the predefined cluster centers to generate superpixel. 
Yang \emph{et al.} designed a fully-connected convolutional network (S-FCN)\cite{S-FCN} that adopts an encoder-decoder structure, which simplifies the iteratively clustering step of SSN by assigning each pixel into the 9-neighbor grid. Though S-FCN improves the segmentation efficiency, it is still supervised by the segmentation labels, and needs upsampling the input images to generate large number of superpixel.
Recently, Suzuki utilized the CNN to unsupervisely generate superpixel with regular information maximization (RIM)\cite{RIM}. It trains a randomly initialized CNN to reconstruct the input image while minimizing the entropy among each superpixel. However, it needs to reinitialize the parameters of the network and takes a long time to reach convergence when generating superpixel for each image.

\section{Method}
\begin{figure*}
\centering
\includegraphics[width=0.95\textwidth]{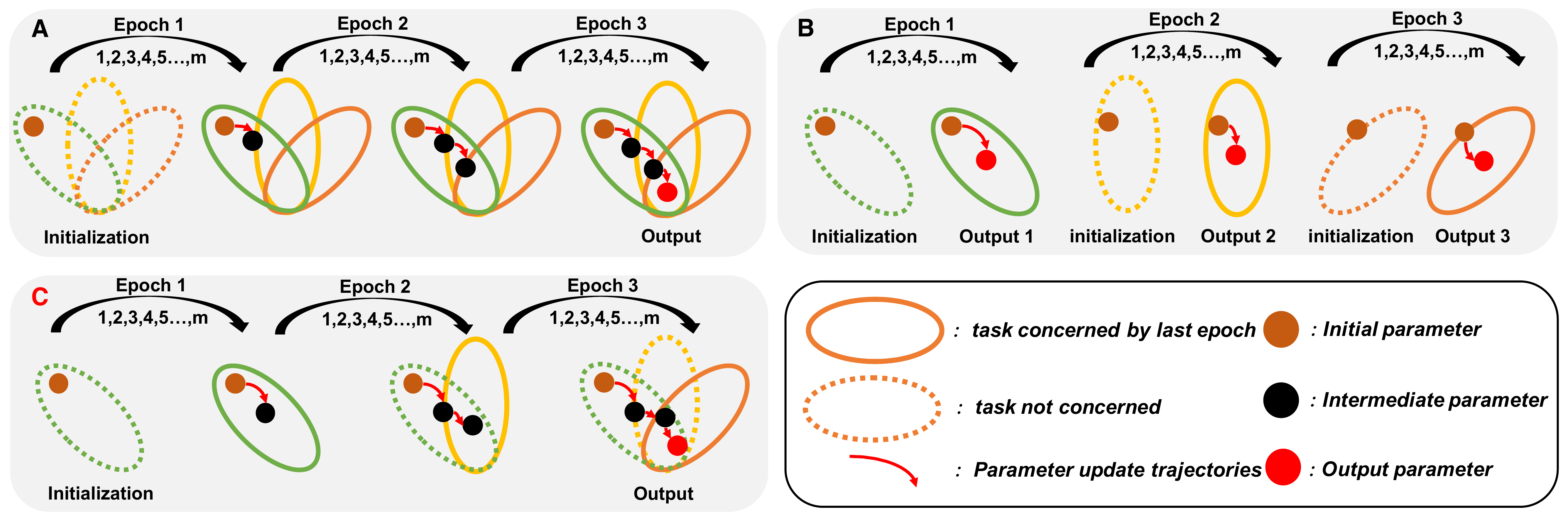}
\caption{The training strategy of our LSN-Net and other learning-based superpixel segmentation methods. ``Ellipses" with different colors mean different clustering tasks (images). Each ``dot" means the parameters of the network during training progress. \textbf{ \sf{A}.} The multi-task learning strategy, which learns \emph{a unified embedding} by optimizing \emph{the whole task set}. \textbf{ \sf{B}.} The isolated learning strategy, which respectively learns \emph{a unique embedding} for \emph{each task}. \textbf{ \sf{C}.} The lifelong learning strategy of our proposed LNS-Net, which learns \emph{a unified embedding} by separately optimizing \emph{each task}.}
\label{fig:intro}
\end{figure*}
In this section, we begin with defining the superpixel segmentation as the lifelong learning task, where the segmentation process of each image can be viewed as an independent clustering task. Then, we propose a convolutional network structure called LNS-Net which contains: 1) feature embedder module (FEM); 2) non-iterative clustering module (NCM); 3) gradient rescaling module (GRM). Finally, we give our loss function, which does not require any manual labels to supervise training process.

\subsection{Problem Definition}

In general, the learning strategy of existing CNN-based superpixel segmentation methods can be divided into two categories. One is the multi-task learning strategy\cite{SSN, S-FCN}, which learns a unified embedding based on the whole image set depicted in Figs.~\ref{fig:intro} \textcolor{red}{A}. It updates the weight parameter based on all images during the whole training process which requires large amounts of computation resources. The other is the isolated learning strategy\cite{RIM}, which respectively learns a unique embedding for each image as shown in Figs.~\ref{fig:intro} \textcolor{red}{B}. Though this strategy does not require to maintain all the images, a unique parameter space needs to be repeatedly found for each image, which is time-consuming and cannot generalize to other images. In order to overcome these drawbacks, our proposed LNS-Net sequentially refines the unified embedding based on a certain image, which is a classic case of lifelong learning. As shown in Figs.~\ref{fig:intro} \textcolor{red}{C}, our lifelong learning strategy only focuses on one image per epoch and intends to maintain the performance for the images learned in prior epoches simultaneously.

To theoretically define our sequential learning strategy, we start with the segmentation of a specific image $I$, which aims to segment the image $I$ into $K$ compact regions by assigning the label for each pixel of the entire image map $\bm{L}^{I}$.  It can be seen as a clustering task $T^{I}$ where each pixel $i$ with feature $\bm{x}_{i} = \{r, g, b, p_{x}, p_{y} \}$ forms the samples $\bm{X}^{I} = \{\bm{x}_1, \bm{x}_2, ..., \bm{x}_N\}$. Supposing the index set of the cluster centers is $\bm{S}^{c}$, a following cluster step $\bm{L}^{I} = \bm{c}(\bm{Z}^{I} | \bm{S}^{c})$ is used to generate the label for each pixel, where $\bm{c}(\cdot)$ is a cluster function. $\bm{Z}^{I} = \bm{e}(\bm{X}^{I} | \mathbf{W}_e)$ is a learned embedding map to project the samples $\bm{X}^{I}$ into a clustering-friendly space with function $\bm{e}(\cdot)$. The learning weight $\mathbf{W}_e$ can be optimized by $\mathbf{W}_e = \mathbf{W}_e - \alpha * \mathrm{d}\mathbf{W}_e$ with $\mathrm{d}\mathbf{W}_e = \frac{\partial \mathcal{L}}{\partial \mathbf{W}_e}$, where $\mathcal{L}$ is the loss function and $\alpha$ is the learning rate.

Further, assuming that we have a set of images $\mathbb{I} = \{I_1, I_2, ..., I_n \}$, the segmentation of $\mathbb{I}$ can be modeled as a series of clustering tasks $\mathbb{T} = \{T^1, T^2, ..., T^n\}$. Different from the existing models that either obtain the embedding $\bm{e}(\bm{X} | \mathbf{W}_e)$ by optimizing $\mathbf{W}_e$ based on the whole set $\mathbb{T}$\cite{SSN, S-FCN} or separately training an embedding $\bm{e}^{i}(\bm{X}^{i} | \mathbf{W}^{i}_{e})$ for each task $T^{i}$\cite{RIM} to obtain the cluster-friendly feature $\bm{Z}$, we aim at optimizing each task $T^{i}$ separately to generate a unified embedding function $\bm{e}(\bm{X}^{i} | \mathbf{W}_{e})$. During the optimization, the retentivity of $\mathbf{W}_{e}$ for prior tasks is also enhanced by a scaling function $\bm{\psi}(\mathrm{d}\mathbf{W})$. Finally, with the cluster-friendly features $\bm{Z}$, pixels can be labeled by the cluster function $\bm{L} = \bm{c}(\bm{Z} | \bm{S})$ with optimal seed nodes $\bm{S}$.

\begin{figure*}
\centering
\includegraphics[width=0.98\textwidth]{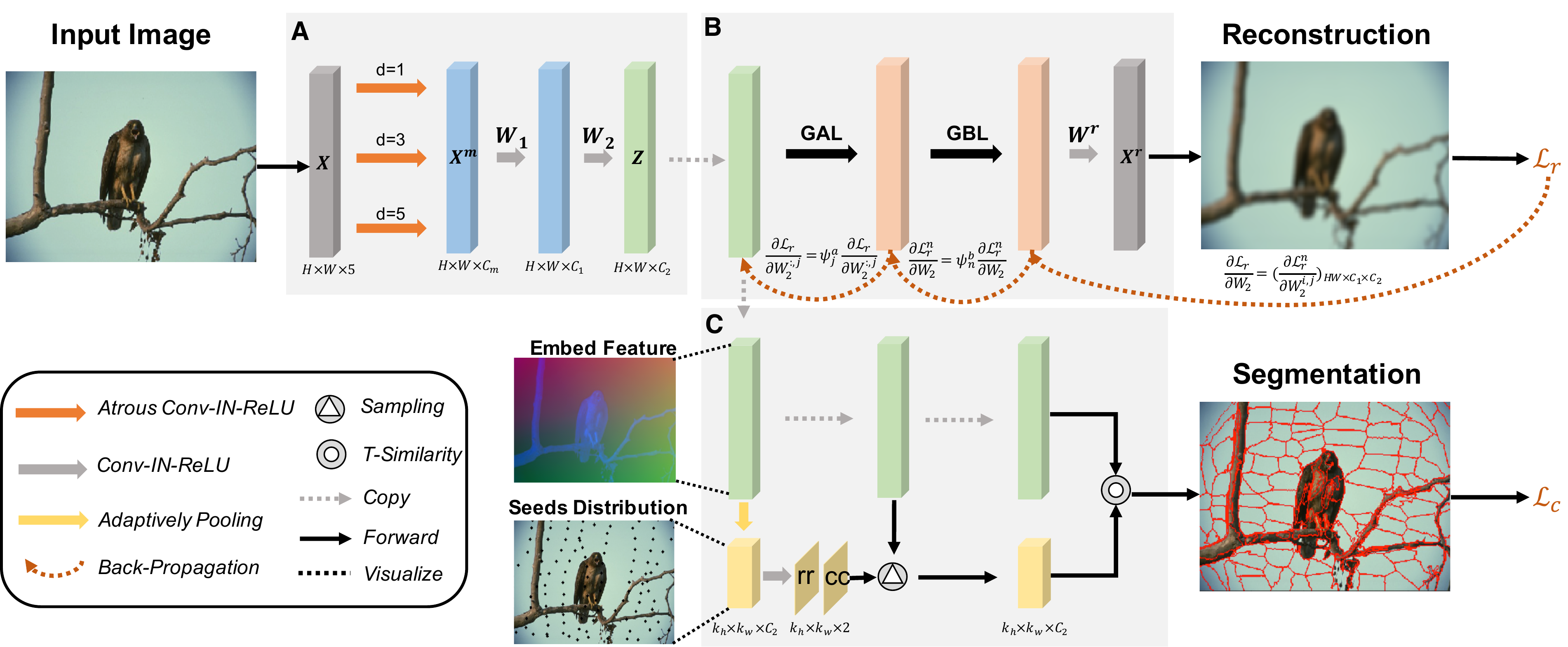}
\caption{The network structure of our LNS-Net. ``Seeds distribution" visualizes the seed node generated by NCM. ``Embed feature" visualizes the cluster-friendly feature map $\bm{Z}$ with the help of PCA dimension reduction. \textbf{ \sf{A}.} The structure of our FEM. \textbf{ \sf{B}}. The structure of our GRM. \textbf{ \sf{C}.} The structure of our NCM.}
\label{fig:structure}
\end{figure*}
\begin{algorithm} 
\small
\caption{Flow of the proposed LNS-Net} 
\label{alg1} 
\begin{algorithmic}[1] 
\REQUIRE Cluster tasks $T$, Feature set $X$, Max epoches $M$
\STATE Initialize the learning rate and parameters
\FOR{$T^{i}$ in $\mathbb{T}$} 
\STATE Select the pixels feature $\bm{X}^i$, Set $m = 0$
	\FOR{$m < M$}
		\STATE Get the $\bm{Z}^{i} = \bm{e}(\bm{X}^{i} | \mathbf{W}_{e})$ by FEM.
		\STATE Get the labels $\bm{L} = \bm{c}(\bm{Z}^{i} | \bm{S})$ by NCM.
		\STATE Rescale the gradient: $d\mathbf{W}_e = \bm{\psi}(d\mathbf{W}_e)$ by GRM. 
		\STATE Backward update $\mathbf{W}_{e}$ and other parameters.
	\ENDFOR
\ENDFOR
\end{algorithmic} 
\end{algorithm}

The flow of our LNS-Net is given in Algorithm 1. We separately optimize each clustering task $T^{i}$ and train a network that contains three proposed modules to implement the functions $\bm{e}(\cdot)$,  $\bm{c}(\cdot)$ and $\bm{\psi}(\cdot)$ respectively. Once $T^{i}$ has been optimized, we start to focus on the next task $T_{i+1}$ until all tasks are trained.

\subsection{Network Design}
\label{network_design}
The structure of proposed LNS-Net shown in Fig.~\ref{fig:structure} contains three parts: 1) the proposed lightweight FEM embeds the original feature into a cluster-friendly space; 2) the proposed NCM assigns the label for pixels with the help of a seed estimation module, which automatically estimates the indexes of seed nodes; 3) the proposed GRM adaptively rescales the gradient for each weight parameter based on the channel and spatial context to avoid catastrophic forgetting for the sequential learning.


\noindent \textbf{Feature Embedder Module: } Actually, superpixel segmentation is based on the low-level color and spatial features rather than the high-level semantic features. We argue that the feature embedders used by other CNN-based methods\cite{SSN, S-FCN, RIM} are too redundant for the superpixel segmentation, due to their large number of channels and receptive field. As a alternative, our FEM only uses two convolution layers with an additional atrous spatial pyramid pooling (ASPP)\cite{ASPP} to enlarge the receptive field rather than go deeper with spatial pooling, which can better preserve details with fewer parameters. As shown in Fig.~\ref{fig:structure} \textcolor{red}{A}, the LAB (or RGB) features and the position indexes of pixels are concatenated and fed into the ASPP structure to capture multi-scale information:
\begin{equation}
	\bm{X}^m = \sigma( concat(\bm{X} * \mathbf{H}^0, \bm{X} * \mathbf{H}^1, \bm{X} * \mathbf{H}^2 ) )
\label{eq:1}
\end{equation}
\noindent where ``$*$" is the convolutional operator, $\bm{X} \in \mathbb{R}^{N \times 5}$ is the input feature and $\bm{X}^m \in \mathbb{R}^{N \times C_m}$ is the multi-scale feature. $\mathbf{H}^d \in  \mathbb{R}^{5 \times \frac{C_m}{3}}$ is the convolution with dilation range $d$, $\sigma$ is the non-linear function implemented by ReLU. Then two $3 \times 3$ convolution are used to embed the multi-scale feature $\bm{X}^m$ into the cluster-friendly space:
\begin{equation}
	\bm{Z} = \sigma( \sigma(\bm{X}^m * \mathbf{W}_{1}) * \mathbf{W}_{2}) )
\label{eq:2}
\end{equation}
\noindent where $\bm{Z} \in \mathbb{R}^{N \times C_2} $ is the cluster-friendly feature, $\mathbf{W}_1 \in \mathbb{R}^{C_m \times C_1}, \mathbf{W}_2 \in \mathbb{R}^{C_1 \times C_2}$ are the parameter matrixes. 


\noindent \textbf{Non-iterative Clustering Module: } Once the embedding feature $\bm{Z}$ has been obtained, the superpixel can be generated by clustering the features in $\bm{Z}$ with the initialized cluster centers $\bm{S}^{c}$. However, those cluster centers usually have a coarse distribution such as the center of grid. The cluster step with time complexity $O(N*K)$ needs to iteratively refine the distribution of the centers. Such refining process is unintegrable in majority cases. Though the recent work\cite{SSN} makes it integrable, it still needs nearly 10 iterators to reach convergence. To avert this time-consuming process, our NCM uses a seed estimation layer (SEL) to estimate a satisfactory cluster center distribution based on $\bm{Z}$ by learning the offsets to shift those coarse centers into a more reasonable distribution, \emph{i.e.} the seed nodes $\bm{S}$.

As shown in Fig.~\ref{fig:structure} \textcolor{red}{C}, $\bm{Z}$ is adaptively pooled into a low-resolution $\bm{Z}^{k} \in \mathbb{R}^{K \times C_2}$ , where $K$ is the number of target superpixel. Then, a linear project with sigmoid activation is used to learn the offsets contained by a two-dimension vectors $\bm{F}_{i} \in \mathbb{R}^{K \times 2}$:
\begin{equation}
	\bm{F} = sigmoid(\bm{Z}^{k} * \mathbf{W}_s)
\label{eq:3}
\end{equation}
\noindent where $\mathbf{W}_s \in \mathbb{R}^{C_2 \times 2}$ is the parameter matrix of the linear project, which can be learned by Adam\cite{Adam}. The two dimensions of $\bm{F}=\{\bm{rr}, \bm{cc}\}$ can be viewed as the crosswise shift ratio $\bm{rr}$ and the longitudinal shift ratio $\bm{cc}$. Then, we restrict their shift scopes inside the corresponding grid by:
\begin{equation}
	\begin{split}
		\triangle \bm{r} = (\bm{rr} - 0.5) * R , ~~~~~~ \triangle \bm{c} = (\bm{cc} - 0.5) * C \\
	\end{split}
\label{eq:4}
\end{equation}
\noindent where $R$, $C$ are the number of rows and columns of the image, respectively. Next, the offsets $(\triangle \bm{r}, \triangle \bm{c})$ are added on the corresponding center to obtain the optimized seed nodes $\bm{S}$:
\begin{equation}
	\bm{S} = \bm{S}^{c} + (\triangle \bm{r} * R + \triangle \bm{c})
\label{eq:5}
\end{equation}
\noindent where $\bm{S}^{c}$ is the coarse clustering center indexes and $\bm{S}$ is the output seed node indexes.

Finally, the clustering layer (CL) of NCM is used to assign the labels $\bm{L}$ for pixels based on $\bm{S}$. The CL firstly adopts the $t$-distribution kernel to measure the similarity between each pixel and seed node:
\begin{equation}
	\bm{P}_{ik} = \frac{(1 + || \bm{Z}_{i} - \bm{Z}_{\bm{S}_{k}} ||^{2}) ^ {- \frac{1}{2}}} {\sum_{k}{(1 + || \bm{Z}_{i} - \bm{Z}_{\bm{S}_{k}} ||^{2}) ^ {- \frac{1}{2}}}}
\label{eq:6}
\end{equation}
\noindent where $\bm{P} = \in \mathbb{R}^{N * K}$ is the soft assignment between each pixel and seed node. Finally, the label of each pixel can be obtained by selecting the seed with maximal similarity:
\begin{equation}
	\bm{L}_{i} = argmax_{k}(\bm{P}_{i0}, \bm{P}_{i1}, ..., \bm{P}_{ik})
\label{eq:7}
\end{equation}
%
\noindent \textbf{Gradient Rescaling Module:} Considering the images are sequential in our learning step, the network will face the catastrophic forgetting that the network over-fits the domain of current task without concerning prior tasks. To overcome this problem, our proposed GRM serves two purposes: 1) using the proposed gradient adaptive layer (GAL) to perceive the importance for the gradient on different \textit{feature channels} to avoid over-fitting; 2) using the proposed gradient bi-direction layer (GBL) to generate confrontation based on the \textit{spatial context} to improve  generalizability. 

Specifically, both GAL and GBL are backed by a reconstruction head that reconstructs the clustering-friendly feature into the original input features (both spatial and color features) with an additional linear project:
\begin{equation}
	\bm{X}^{r} = \bm{Z} * \mathbf{W}^{r}
\end{equation}
\noindent where $\bm{X}^{r} \in \mathbb{R}^{N \times 5}$ is the reconstruction feature whose first three columns are the color features (RGB/LAB) and the rest are spatial features (position indexes), which is respectively supervised by the reconstruction loss $\mathcal{L}_{r}$. $\mathbf{W}^{r} \in \mathbb{R}^{C_2 \times 5}$ is the parameter for the linear project.

Based on the reconstruction head, the mean reconstruction strength $\bm{g}(\mathbf{W}^r)$ can be defined to represent the importance for the channel of the cluster-friendly feature: 
\begin{equation}
	\bm{g}(\mathbf{W}^r) = (\sum_{i=1, 2, 3} |\mathbf{W}^r_{:, i} / 3|) \odot (\sum_{i=4, 5}|\mathbf{W}^r_{:, i} / 2|)^T
\end{equation}
\noindent where $\odot$ is the Hadamard product. The higher $\bm{g}(\mathbf{W}^r)_{:, c}$ is, the more $\bm{Z}_{:, c}$ contributes for reconstructing $\bm{X}$ in forward-propagation, \emph{i.e.} this channel has already better fit the domain of current task. Thus, even though $\bm{g}(\mathbf{W}^r)_{:, c}$ drops in the following tasks, which causes a high gradient $\mathrm{d}\mathbf{W}_{:, c}$, this weight $\mathbf{W}^r_{:, c}$ should be maintained to avoid over-fitting. To achieve this, a vector $\bm{m} \in \mathbb{R}^{1 \times C_2}$ is defined to preserve the historical $\bm{g}(\mathbf{W}^r)_{:, c}$, which is initialized as an all-one tensor and progressively updated during the sequential training step:
\begin{equation}
	\bm{m} = \lambda * \bm{g}(\mathbf{W}^r) + (1-\lambda) * \bm{m} \\
\end{equation}
\noindent where $\lambda$ is to adjust current and history gradient scale. 

Based on $\bm{m}$, our GAL is designed to rescale the gradient of the weight parameter in FEM to avoid overfitting, which works as a ``pseudo-function" $\bm{R}^{a}(\cdot)$ with the following forward- and back-propagation:
\begin{equation}
	\begin{split}
		& \bm{R}^{a}(\bm{X}_{n, :}) = \bm{X}_{n, :} * \bm{I} \\
		& \frac{\mathrm{d}\bm{R}^{a}}{\mathrm{d}\bm{X}_{n, :}} = \bm{\psi}^{a} * \bm{I} = \frac{\bm{g}(\mathbf{W}^r)}{\bm{g}(\mathbf{W}^r) + \bm{m}} * \bm{I}
	\end{split}
\label{eq:AGR}
\end{equation}
\noindent where $\bm{I} \in \mathbb{R}^{C_2 \times C_2}$ is the identity matrix. In the forward-propagation, GAL acts as an identity transform which percepts the importance for each channel by $\bm{g}(\mathbf{W}^r)$ to preserve the historical memory matrix $\bm{m}$. During back propagation, GAL scales the gradient of the weight parameters, which lowers the gradient of weights corresponding to the channel with high $\bm{m}_{c}$ to avoid over-fitting the current task.

Though the proposed GAL can avoid over-fitting by concerning the historical strength of the channel, it treats each pixel equally without considering their spatial context. Actually, the superpixel segmentation is a dense prediction task, which aims to balance the contour adherence and spatial compactness. This requires the model biasing the color features for pixels near contours, while concerning both the color and the spatial features for pixels in smooth areas. To compensate this, GBL is proposed to rescale the gradient based on the spatial context to avoid overfitting. It  generates bi-direction gradient scale based on the contour map $\bm{B}$ to confound the reconstruction strength for the spatial features of the pixels near contours. The forward- and back-propagation of our GBL are:
 \begin{equation}
	\begin{split}
		& \bm{R}^{b}(\bm{X}_{n, c}) = \bm{X}_{n, c} \\
		& \frac{\mathrm{d}\bm{R}^{b}}{\mathrm{d}\bm{X}_{n, c}} = \bm{\psi}^{b}_{n} = \left\{
			\begin{array}{rcl}
				 1      &   ,   & {\bm{B}_n      \le       \epsilon}\\
				- \bm{B}_n       &  ,    & {\bm{B}_n      >      \epsilon}
			\end{array} \right.
	\end{split}
\label{eq:GBL}
\end{equation}
\noindent where $\bm{R}^{b}(\cdot)$ is the ``pseudo-function" of our GBL. In the forward step the GBL also acts as an identity map. While in the backward step, the GBL generates a bi-direction gradients for the different pixels $i$ based on their contour map $\bm{B}_{i}$, which makes the pixels near contours bias the color feature reconstruction even though having a confounding spatial information to enhance generalizability. 
 
\subsection{Loss Function}

A two-terms loss is used to supervise the sequential training step for our network, which can be formulated as:
\begin{equation}
	\mathcal{L} = \mathcal{L}_{c} + \beta*\mathcal{L}_{r}
\end{equation}
\noindent where $\mathcal{L}_{c}$ is the clustering loss that encourages the network to group pixels with similar probability. $\mathcal{L}_{r}$ is the reconstruction loss to help the cluster-friendly feature $\bm{Z}$ concern both color details and spatial information. $\beta$ is used to balance the two losses. 

\noindent \textbf{Cluster Loss:}  We propose a range-limited cluster loss to train our network without requiring manual label. It can be formulated as a regularized KL divergence between the limited range soft assignment $\widetilde{\bm{P}}$ with its reference distribution $\widetilde{\bm{Q}}$:
\begin{equation}
	\mathcal{L}_{c} = \sum_{i} \sum_{k} \widetilde{\bm{Q}}_{ik} log \widetilde{\bm{Q}}_{ik} - \widetilde{\bm{Q}}_{ik} log \widetilde{\bm{P}}_{ik}  + l(\bm{P})
\end{equation}
\noindent where $l(\cdot)$ is a regular term. The limited range soft assignment $\widetilde{\bm{P}}$ enhances the probability for allotting pixels into its ``Top\text{-}$n$" nearest seed nodes, which improves the compactness of the segmentation result. Specifically, the spatial distance $\bm{D}_{ik}$ between the pixel $i$ and the seed node $k$ is firstly calculated based on the $l_1$ distance on their spatial indexes. Then, we define $\mathbb{V}_{i} = \text{Top-}n_{k}(\bm{D}_{i0}, \bm{D}_{i1}, ..., \bm{D}_{ik})$ as the ``Top\text{-}$n$" seeds set for pixel $i$ and use it to build a mask matrix, which masks the elements between the pixel-seed pairs with large distance:
\begin{equation}
	\bm{M}_{ik}=\left\{
		\begin{aligned}
			0 &,~~k \in \mathbb{V}_{i} \\
			1 &,~~k \notin \mathbb{V}_{i}
		\end{aligned}
		\right.
\end{equation}
\noindent Finally, the limited range soft assignment $\widetilde{\bm{P}} = \bm{P} \odot \bm{M}$ can be obtained by adding masks on the original assignment $\bm{P}$, \emph{i.e.}, where $\odot$ is the Hadamard product.

To improve the cluster purity and penalize the superpixel with too large size, we follow Xie \emph{et al.}\cite{DEC} and define $\widetilde{\bm{Q}}$ without requiring the manual labels:
\begin{equation}
	\widetilde{\bm{Q}}_{ik} = \frac{\widetilde{\bm{P}}_{ik}^2 / \sum_{i} \widetilde{\bm{P}}_{ik}} {\sum_{j} (\widetilde{\bm{P}}_{ik}^2 / \sum_{i} \widetilde{\bm{P}}_{ik}) }
\end{equation}
A regularized term is also added to avoid the local optimum where pixels are assigned into the seed node that not in $\mathbb{V}_{i}$:
\begin{equation}
	l(\bm{P}) = \frac{\bm{P} \odot \bm{M}} {\bm{P} \odot (\bm{1} - \bm{M})}
\end{equation}
\noindent \textbf{Reconstruction Loss:}  Reconstruction loss is a crucial part for our proposed GRM to rescale the gradient of weight parameter. As discussed in Sec.\ref{network_design}, $\mathcal{L}_{r}$ supervises both reconstruction of the input color and spatial features, which can be define as:
\begin{equation}
	\mathcal{L}_{r} = \mathcal{L}_{r_{c}} + \phi * \mathcal{L}_{r_{s}}
\end{equation}
\noindent $\mathcal{L}_{r_{c}}$ is the reconstruction loss of color feature,  $\mathcal{L}_{r_{s}}$ is the reconstruction loss of spatial feature and $\phi$ controls the trade-off between $\mathcal{L}^{i}_{r_{c}}$ and $\mathcal{L}^{i}_{r_{s}}$. Specifically, MSELoss between the reconstruction result and original input is used as the reconstruction loss for $\mathcal{L}_{r_{c}}$ and $\mathcal{L}_{r_{s}}$.

From another view, due to the bi-direction gradient generated by our GBL, the reconstruction loss for our network is also equivalent to :
\begin{equation}
 	\mathcal{L}_{r} = \sum_{i \notin \mathbb{V}_{b}} (\mathcal{L}^{i}_{r_{c}} + \phi * \mathcal{L}^{i}_{r_{s}}) + \sum_{i \in \mathbb{V}_{b}} (\mathcal{L}^{i}_{r_{c}} - \bm{B}_i * \phi * \mathcal{L}^{i}_{r_{s}})
\label{eq:loss_sb}
\end{equation}
 \noindent where $\mathbb{V}_{b}=\{n | \bm{B}_n >  \epsilon\}$ is the counter pixel set. In Eq.~(\ref{eq:loss_sb}), the spatial reconstruction part for pixels near contours, \emph{i.e.} $\sum_{i \in \mathbb{V}_{b}}(\bm{B}_i * \phi * \mathcal{L}^{i}_{r_{s}})$, serves as a regularization term that avoids the cluster-friendly feature map $\bm{Z}$ paying much attention on the spatial feature for the pixels in $\mathbb{V}_{b}$.

\begin{figure*}
\centering
\subfigure[Ablation Experiments Result]{
\includegraphics[width=0.485\textwidth]{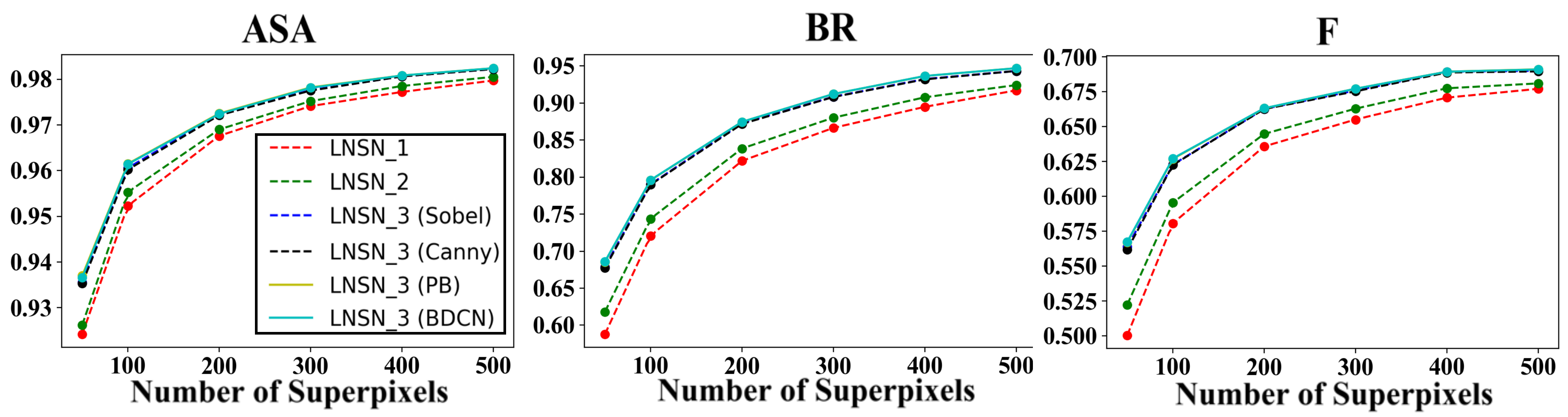}
\label{Ablation}}
\subfigure[BSDS dataset]{
\includegraphics[width=0.485\textwidth]{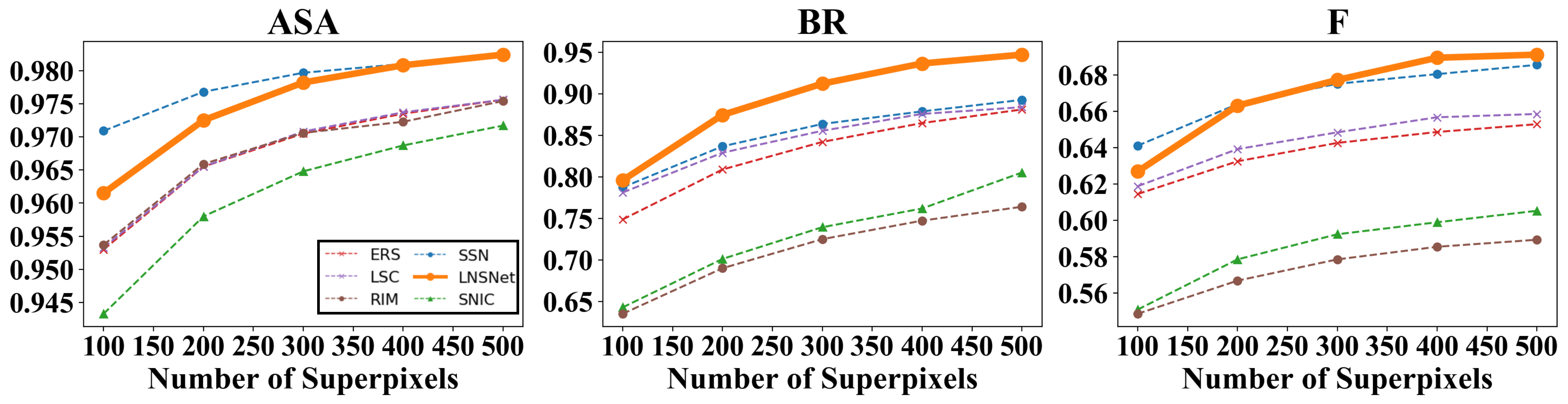}
\label{BSDS}}
\subfigure[DRIVE dataset]{
\includegraphics[width=0.485\textwidth]{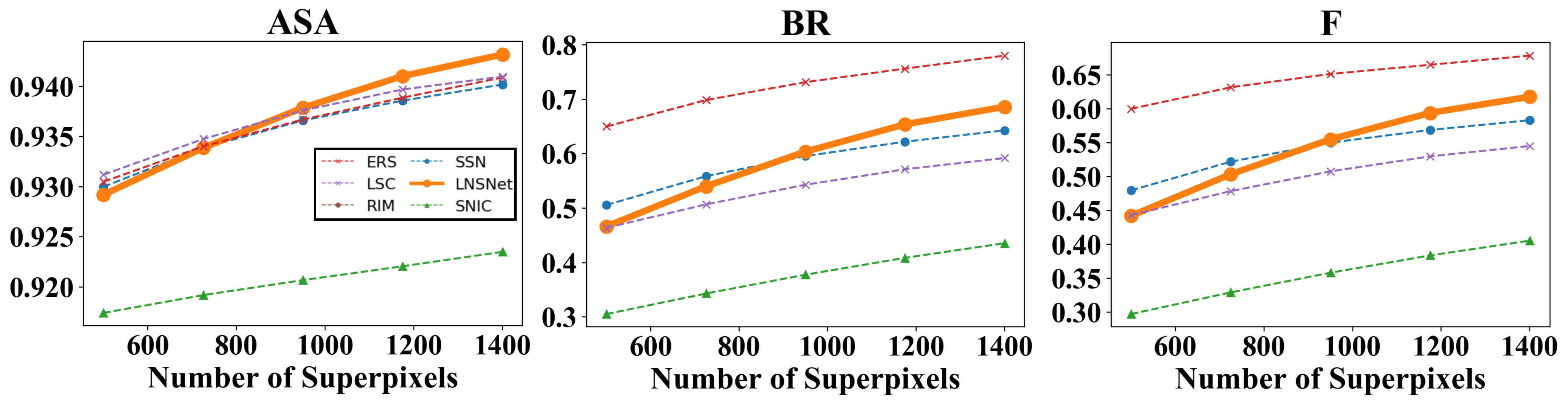}
\label{DRIVE}}
\subfigure[DME dataset]{
\includegraphics[width=0.485\textwidth]{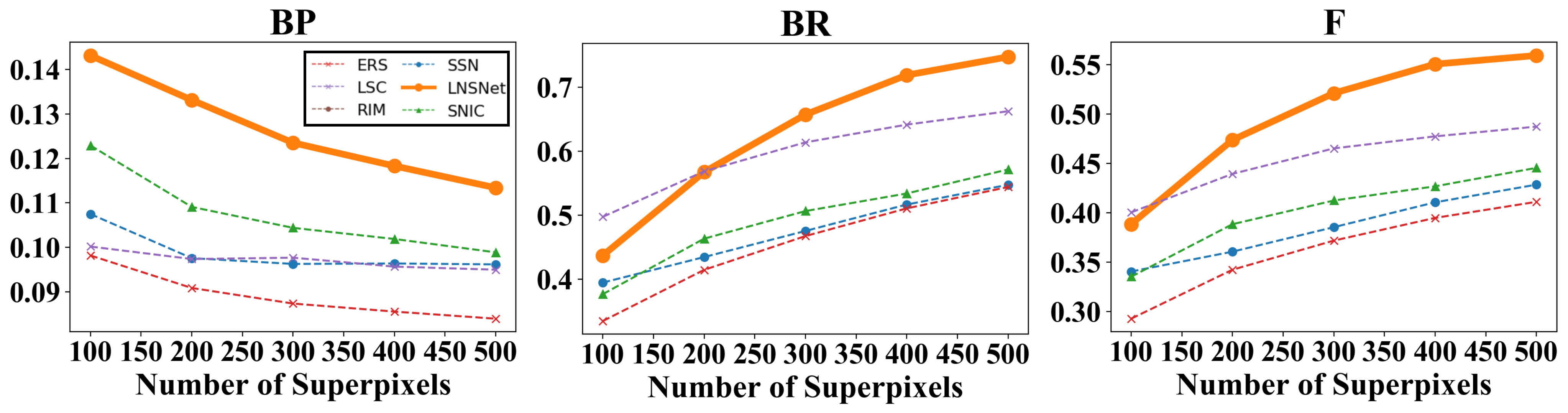}
\label{DME}}
\caption{The experimental results for ablation strudies of the proposed LNS-Net and the comparison for different superpixel segmentation methods on the BSDS, DRIVE, DME datasets. Better view in color and zoom in four times.}
\end{figure*}
\section{Experiment}

We conduct experiments on three datasets to demonstrate the effectiveness of the proposed model. We firstly introduce the settings of our experiment including the implementation details of our LNS-Net and the evaluation metrics. Then, ablation studies are performed on BSDS500 dataset to evaluate the different modules of our LNS-Net. Finally, we compare our proposed LNS-Net with other superpixel segmentation methods.
\subsection{Settings}
\noindent \textbf{Implementation Details:} Our LNS-Net is implemented with PyTorch. The numbers of the three channels in FEM are set as $C_m = 10$, $C_1 = 10$, $C_2 = 20$. For the loss function, we set the balance parameter $\beta$, $\phi$ and the neighbor number $n$ as $10$, $1$, $9$ respectively. During the sequential training step, each image is sequentially trained 50 epoches where the first 40 epoches focus on feature learning so $\mathbf{W}_s$ of the seed estimator layer is locked. The last 10 epoches serve as updating the seed distribution, where all the weights of FEM are locked. Adam \cite{Adam} with learning rate $0.0003$ is used to optimize the parameters. Note that our training step do not require any manual label. And in the test step, only our FEM and NCM are used to generate superpixel efficiently.

\noindent \textbf{Evaluation Metrics:} In our experiments, the Boundary Recall (BR), the boundary Precision (BP), the Achievable Segmentation Accuracy (ASA) and the F-beta Score (F) are used to evaluate the superpixel segmentation. Considering that the recall is more important than the precision for superpixel segmentation, beta is set $4$ for F-beta. For the dataset that has more than one groundtruth such as BSDS500\cite{BSDS}, we choose the best one among all the ground truths as the listed score. Moreover, like SSN\cite{SSN} and RIM\cite{RIM}, we also use the same strategy to enforce the spatial connectivity before calculating the evaluation metric.
\subsection{Ablation Study}
Ablation studies are conducted on the BSDS500 dataset\cite{BSDS} to show the effectiveness of the proposed modules. We explain three type LNS-Net structures in details: \textit{LNS-Net$_1$} is the most simple backbone, which only uses FEM to embed the feature and cluster the pixels with grid seed node; \textit{LNS-Net$_2$} adds SEL of our proposed NCM to automatically generate seed nodes; \textit{LNS-Net$_3$} further adds GRM to adaptively rescale the gradient of weight parameter based on reconstruction results to avoid over-fitting by concerning feature channel and spatial context. Note that, the contour map generated by both unsupervised learned methods (Sobel, Canny\cite{Canny}) and supervised learned methods (PB\cite{PB}, BDCN\cite{BDCN}) are also tested for the proposed GRM.

The performance of these models is shown in Fig.~\ref{Ablation}. It can be seen that, even using the simple grid seed nodes (LNS-Net$_1$), our model outperforms the unsupervised method RIM\cite{RIM} by a large margin profited by the cluster-friendly feature space generated by our FEM. While, adding the proposed seed estimation layer to automatically generate seed nodes (LNS-Net$_2$), the BR, ASA, F are further improved facilitated by the more suitable seed distribution. Next, when the proposed GRM is added (LNS-Net$_3$), overfitting is avoided, bringing an obvious improvement in the four evaluation criteria. Moreover, it can be also seen that unsupervised contour (dotted line) are comparable to the supervised-learned contours (full line), which means our GRM is not sensitive to different contour priors.

%
\begin{figure*}
\centering
\includegraphics[width=0.92\textwidth]{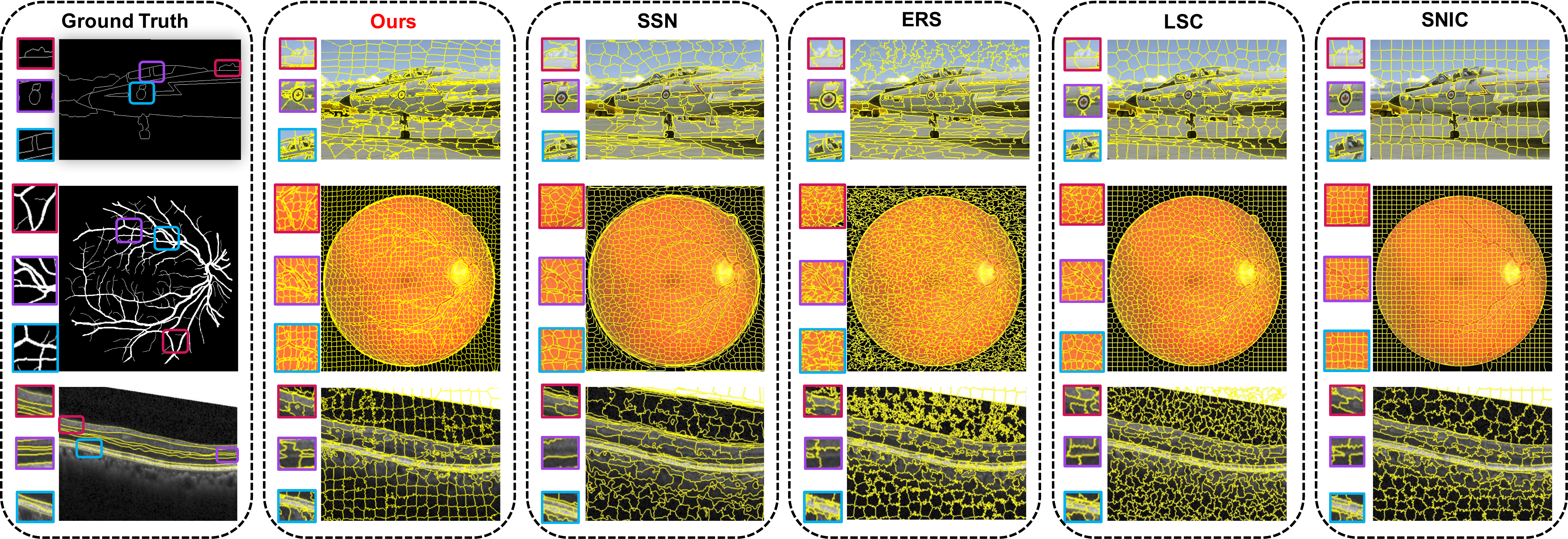}
\caption{The quantitative results for different superpixel segmentation methods on the BSDS dataset (top row), DRIVE dataset (middle row) and DME dataset (bottom row). Better view in color and zoom in four times. }
\label{fig:vis}
\end{figure*}
\subsection{Results}
In this section, three datasets from different domains are used to compare the performance of our proposed LNS-Net with other superpixel segmentation algorithms, including the graph-based ERS\cite{ERS}, LSC\cite{LSC}, the gradient-based SNIC\cite{SNIC} and the CNN-based SSN\cite{SSN}, RIM\cite{RIM}. Visualization of the segmentations results of these methods on the three datasets are shown in Fig.~\ref{fig:vis}. 

\noindent \textbf{BSDS500 dataset\cite{BSDS}} is the standard benchmark for superpixel segmentation which contains 200 training images, 100 validation images and 200 test images. The size of image in this dataset is $481 \times 321$. Each image has more than 5 segmentation ground truths labeled by different person. Thus, we choose one of the ground truth that can achieve the highest segmentation scores in this study. Considering that SSN\cite{SSN} is a supervised superpixel segmentation method that needs training the model on the training set and validation set to optimize the parameters, we only compare the performance on the test set for all superpixel segmentation method mentioned above. Quantitative results on BSDS dataset are shown in Fig.~\ref{BSDS}, it can be seen that our LNS-Net has the highest performance among all the unsupervised superpixel segmentation methods (ERS\cite{ERS}, SNIC\cite{SNIC}, LSC\cite{LSC}, RIM\cite{RIM}). This benefits from our sequential training strategy, which can unsupervisely optimize the model parameters. Moreover, our LNS-Net is more sensitive to the contours in a broad sense rather than only the semantic boundaries as shown in Fig.~\ref{fig:vis}. This trait contributes to our higher BR than the supervised segmentation method SSN.

\noindent \textbf{DRIVE dataset\cite{DRIVE}} is consisted of 40 retinal images with size $565 \times 584$ and the corresponding ground truth for their blood vessel. The domain of DRIVE dataset are very different from the images of BSDS500 as shown in Fig.~\ref{fig:vis}, thus the same models trained on BSDS500 from each learning-based method are used to test their generalizability. Experimental results of the different methods on these 40 retinal images are listed in Fig.~\ref{DRIVE}. It can be seen that only graph-based methods ERS\cite{ERS} has higher BR and F than our LNS-Net, because its graph model concerns more global structure of the blood vessel than the other methods. Nevertheless, our LNS-Net is 46 times faster than ERS and has more regular shape of superpixel as shown in Fig.~\ref{fig:vis}. Moreover, our LNS-Net has the highest ASA, indicating that the superpixel generated by our LNS-Net has the highest upper bound for adhering the blood vessel.

\noindent \textbf{Duck DME dataset\cite{DME}} contains 610 B-scans from 10 subjects who have Diabetic Macular Edema (DME). The size of each B-scan is $565 \times 584$ and only 110 of them have manual label for the retina border near their macular. Thus, we select 110 labeled B-scans and crop them into $464 \times 496$ to focus their macular area. For each learning-based method, we also use the same model trained on BSDS500 to segment the three-channel B-scans that expanded from gray scale. Experimental results are shown in Fig.~\ref{DME}, and it can be seen that all scores of our LNS-Net are much more higher than the others, indicating that its effectiveness on catching weak contours for the medical images. Further, the style of B-scans in the DME dataset is also very different from the images in BSDS500 and contain noise as shown in Fig.~\ref{fig:vis}. We can see that the performance of learning-based method SSN deteriorate seriously in the DME dataset, while our LNS-Net can still have a satisfactory result, showing its robust generalizability.

\noindent \textbf{Discussion:} In general, benefiting from the proposed online training step, both the visual impression and the quantitative results demonstrate that our proposed LNS-Net is able to generate better superpixel compared with the unsupervised methods. Even though using an unsupervised sequential training strategy, the superpixel segmentation results generated by LNS-net are still comparable with the supervised learning-based methods. Moreover, LNS-Net has better generalizability with much less complexity \emph{(9 times and 20 times lower in Flops and model size than SSN, respectively)} as shown in Table.~\ref{Time}. 

However, there are still some drawbacks in our LNS-Net, which expected to be addressed in future study. Firstly, due to the sequential training strategy, our model cannot reach complete convergence as the other learning-based methods do. This leads to the existence of trivial regions in the superpixel generated by LNS-Net and needs post-processing to remove them. Secondly, LNS-Net can generate superpixel with relatively regular shapes in the smooth area promoted by the spatial consideration of GBL. But, when facing background with complex texture, the boundary map that assists GBL will contain noises and make the shape of superpixel irregular. Finally, although our LNS-Net uses a lightweight convolutional network and achieves real-time segmentation using GPU, the cluster step still needs to generate distance matrix with $N*K$ dimension, which is inefficient when calculated by CPU with a large $K$.
\begin{table}[!htbp]
\centering
\small
\renewcommand\tabcolsep{3.0pt}
\caption{The performance and complexity of methods for generating 100 superpixel on BSDS dataset with image size $481* 321$}
\label{Time}
\begin{tabular}{ccccccc}
\toprule
~ & Time(ms) & Flops(G) & Size(K) & ASA & Labels & Device\\
\hline
SNIC & $85$ & - & -  & 0.943 & $\times$ & CPU \\
LSC & $269$ & - & - & 0.953 & $\times$ & CPU \\
ERS & $2540$ & - & - & 0.953 & $\times$ & CPU \\
\hline
SSN & $260$ & $13.85$ & $214.5$ & \underline{0.970} & $\checkmark$ & GPU \\
RIM & $34842$ & $64.15$ & $416.14$ &  0.953 & $\times$ & GPU \\
\hline
Ours & \underline{\bm{$55$}} & \underline{\bm{$1.15$}} & \underline{\bm{$11.22$}} & \bm{$0.962$} & \bm{$\times$} & \textbf{GPU} \\
\bottomrule
\end{tabular}
\end{table}
%
\section{Conclusion}
To our best knowledge, this paper is the first work that views superpixel segmentation as a lifelong clustering task. Based on this basis, we propose a CNN-based superpixel segmentation method called LNS-Net. The proposed LNS-Net contains three parts: FEM, NCM, GRM, which is respectively used for feature generation, non-iteratively clustering, and over-fitting avoidance. Experiments show the effectiveness of our LNS-Net in three benchmarks including two medical images datasets. Our method is both efficient and accurate, enabling low latency superpixel generation. 

{\small
\bibliographystyle{ieee_fullname}
\bibliography{egbib}
}

\end{document}